%% file: main.tex
\documentclass[runningheads]{llncs}
\usepackage{array}

  \usepackage{booktabs}
  \usepackage{graphicx}
  \usepackage{amsmath}
  \usepackage{algorithm}
  \usepackage{algorithmic}
  \usepackage{amsfonts}
  \usepackage{subfigure}
  \usepackage{multirow}
  \usepackage{enumitem}
  \usepackage{verbatim}
  \usepackage{xcolor}
  \usepackage{hyperref} 
  \usepackage{makecell}
  \usepackage{cite}
  \usepackage{rotating}
  \usepackage{tabularx}
  \usepackage{adjustbox}
  \usepackage{wrapfig}
  \usepackage{appendix}

\begin{document}
\title{Estimating Individual Dose-Response Curves under Unobserved Confounders from Observational Data}
\titlerunning{Individual Dose-Response Estimation with Hidden Confounders}
%
\author{Shutong Chen\inst{1} \and Yang Li\inst{1}}

\authorrunning{S. Chen, Y. Li}
%
\institute{Shenzhen Key Laboratory of Ubiquitous Data Enabling, Tsinghua Shenzhen International Graduate School, Tsinghua University}
%
\maketitle              
\begin{abstract}
Estimating an individual's potential response to continuously varied treatments is crucial for addressing causal questions across diverse domains, from healthcare to social sciences. However, existing methods are limited either to estimating causal effects of binary treatments, or scenarios where all confounding variables are measurable. In this work, we present ContiVAE, a novel framework for estimating causal effects of continuous treatments, measured by individual dose-response curves, considering the presence of unobserved confounders using observational data. Leveraging a variational auto-encoder with a Tilted Gaussian prior distribution, ContiVAE models the hidden confounders as latent variables, and is able to predict the potential outcome of any treatment level for each individual while effectively capture the heterogeneity among individuals. Experiments on semi-synthetic datasets show that ContiVAE outperforms existing methods by up to 62\%, demonstrating its robustness and flexibility. 
Application on a real-world dataset illustrates its practical utility.

\end{abstract}
\input{1_intro}
\input{2_related_work}
\input{3_formulation}

\input{4_method}
\input{5_simulation_experiment}
\input{6_volunteer}

\section{Conclusions}
Existing methods for estimating unbiased causal effects are restricted either to treatments with binary or discrete values, or scenarios where all confounding variables are measurable. In this work, we propose ContiVAE, a novel framework enabling the estimation of individual dose-response curves in the presence of unobserved confounders. Taking advantage of a variational auto-encoder with a Tilted Gaussian prior distribution as well as an adjusted objective function, ContiVAE demonstrates robustness and flexibility in estimating individual dose-response curves across diverse circumstances, and captures individual differences effectively. In the experiments, we observed consistently the significant improvement of ContiVAE over various benchmark methods. Application to volunteer data also highlights its broader impact in real-world. For the future work, we consider investigating the theory of doubly robust~\cite{kennedy2017nondoublyrobust, bang2005doubly}, 
and refining ContiVAE into a doubly robust estimator for dose-response curves.

%
%
\section{Acknowledgement}
This work is supported in part by the Natural Science Foundation of China (Grant 62371270) and Shenzhen Key Laboratory of Ubiquitous Data Enabling (No.ZDSYS20220527171406015).
%
%
%
\bibliographystyle{splncs04}
\bibliography{main}
%





\input{00_appendix}
\end{document}

%% file: 1_intro.tex
\section{Introduction}
\begin{wrapfigure}[14]{r}{0.3\textwidth}
\vspace{-1.2cm}
    \centering
    \includegraphics[width=1.0\linewidth]{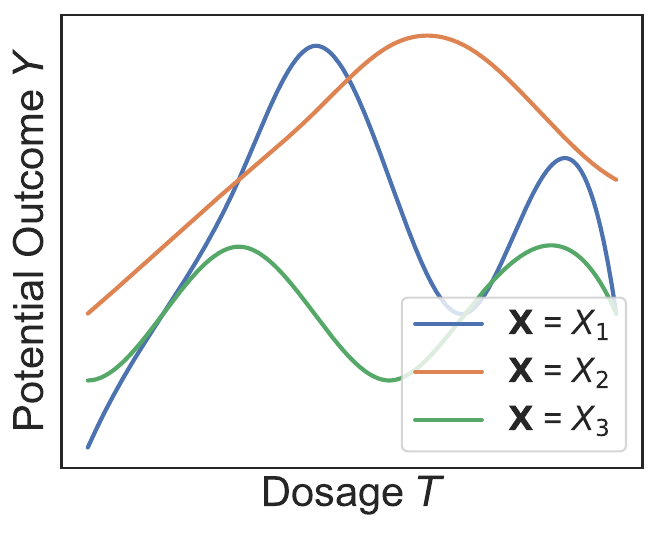}
    \caption{\label{fig: dosecurve} An example of individual dose-response curves. Different colors denote curves of patients with various covariates $\mathbf{X}$ (e.g., age or gender).}
\end{wrapfigure}
While most of the literature and methods in causal inference focus on binary or discrete treatments~\cite{guo2020survey, yao2021survey, tarnet, cevae17, rosenbaum1987ipw, imai2014covariateCBPS, cider2022}, there has been relatively limited exploration in the continuous treatment settings. However, in practice, many causal inquiries involve continuous-valued treatments. For instance, in the medical field, making binary or discrete decision (e.g., whether to adopt a medicine or the type of treatment) is often inadequate, it is also critical to consider continuous-valued decisions, such as the dosage of the medicine, as dosage significantly impacts its effects. Moreover, the effects may vary considerably across patients with different conditions. Therefore, it is important to estimate the individual causal effects of continuous treatments, typically represented as individual dose-response curves (see Figure~\ref{fig: dosecurve}). 

The fundamental problem in estimating causal effects from observational data lies in estimating the \textit{counterfactuals}, i.e., predicting what would happen if a patient received a different treatment, as well as in mitigating the spurious effects caused by \textit{confounders} \textendash \ variables that affects both the treatment and the outcome \textendash\ through properly adjusting for them. The nature of continuous treatments exacerbates the challenges~\cite{scigan2020, schwab2020learningdrnet, fong2018covariateCBGPS}: (1) they involve an infinite number of unobserved counterfactuals, unlike binary treatment scenarios where only one needs to be estimated, and the process of adjusting for confounders becomes more complex and less intuitive; (2) estimating individual dose-response curves requires the counterfactual estimator to effectively capture the individual differences. Furthermore, confounders are not always available in observational data, as factors like social status and living conditions are difficult to quantify or may not be recorded, making them \textit{unobserved confounders}. The presence of unobserved confounders adds complexity to the problem, as not all the confounders can be adjusted for to obtain unbiased causal effects.

To estimate causal effects of continuous treatments, one category of approaches extend methods for discrete treatments, including Generalized Propensity Score (GPS)~\cite{imai2004causalgps} and DRNets~\cite{schwab2020learningdrnet}. 
Another adopts generative networks to predict counterfactual outcomes, such as SCIGAN~\cite{scigan2020}. 
However, such methods rely on the ignorability assumption, which requires the observation and adjustment of all confounders. 
Causal Effects Variational Auto-Encoder (CEVAE)~\cite{cevae17} tackles the challenges of unobserved confounders by leveraging deep latent-variable models to learn them as latent variables and approximately recover causal effects, but its architecture is restricted to binary treatments.

In this work, we aim to address the problem of estimating individual dose-response curves in the presence of unobserved confounders. We introduce ContiVAE, a novel framework that conceives unobserved confounders as latent variables that influence observed data generation. ContiVAE utilizes a variational auto-encoder (VAE) to simultaneously model the distribution of hidden confounders in latent space and their relationships with the observed data (i.e., the covariates $\mathbf{X}$, treatment $T$ and outcome $Y$). We will later show that under specific conditions, individual dose-response curves can be estimated by deriving the joint probability $p(\mathbf{X}, \mathbf{Z}, T, Y)$. 
To effectively capture individual heterogeneity in dose-response curves, we assume a Tilted Gaussian distribution as the prior latent distribution instead of the commonly used Normal distribution~\cite{kingma2013autonormalvae}. 
Unlike the Normal distribution, where the maximum probability density concentrates at a single point, in the Tilted Gaussian distribution, it lies on the surface of a sphere, resulting in an exponentially larger volume 
as the latent dimension increases. Consequently, points in the latent space can be better separated, and individual differences can be represented more effectively. Additionally, latent-variable models for causal inference tend to overly prioritize the reconstruction of high-dimensional or correlated $\mathbf{X}$, leading to incorrect estimation~\cite{rissanen2021criticalcevaefault}. 
To mitigate this, we introduce a scaling parameter $\lambda$ for the reconstruction loss of $\mathbf{X}$, to reduce its contribution to the total loss. 
Our work draws inspiration from CEVAE~\cite{cevae17}, which investigates the application of deep latent-variable models in causal inference for binary treatments, but we make significant modifications to tackle the distinct challenges posed by continuous treatments and mitigate some practical issues such as its tendency to overemphasize covariate reconstruction.
Extensive experimental results show that ContiVAE consistently outperforms several baselines and state-of-the-art models by up to 62\% across various scenarios, demonstrating its robustness and flexibility. Additionally, we apply ContiVAE to a real-world dataset tracking volunteer activities, to estimate the effects of a volunteer's social relationships on her retention. This illustrates how ContiVAE can effectively yield valuable insights to answer real-life causal inquiries.

Our contribution can be summarized as:
\begin{itemize}
    \item We introduce a novel framework, ContiVAE, for estimating individual dose-response curves, filling the gap in identifying causal effects of continuous treatments in the presence of unobserved confounders. 
    \item Extensive experimental results demonstrate that our method achieves state-of-the-art performance in estimating individual dose-response curves across datasets with varying sizes and complexities.
    \item The application of the proposed method to a real-world volunteer dataset shed light on the causal relationship of a volunteer's social network position on retention. 
\end{itemize}

%% file: 2_related_work.tex
\section{Related Work}
\label{sec: relatedwork}

\subsection{Causal Inference for Continuous Treatments}
Many techniques for estimating continuous treatment effects extend propensity score-based methods to continuous settings and estimate the generalized propensity score (GPS)~\cite{imai2004causalgps, fong2018covariateCBGPS}. 
Addtionally, doubly robust frameworks combine propensity score with outcome regression models, mitigating the impact of model misspecification and 
ensuring unbiased estimation of average treatment effects when either of the model is correct.
Recently, neural networks have been increasingly adopted in the field of causal inference due to their non-parametric nature and capacity to model complex functions. TARNET~\cite{tarnet}, a classic framework designed for discrete treatments, is consisted of a shared backbone and multi-task layers trained separately to predict the potential outcome of each treatment value. DRNets~\cite{schwab2020learningdrnet} extends TARNET to continuous treatments by dividing the dosage range of continuous treatments into subintervals and adding a multi-task head for each dosage interval. 
Alternatively, SCIGAN~\cite{scigan2020} utilizes Generative Adversarial Networks (GAN) to model counterfactual distribution. Specifically, it trains a counterfactual generator and a discriminator adversarially, until the generator can produce counterfactual outcomes close enough to the true distribution to fool the discriminator. 

\subsection{Causal Inference with Unobserved Confounders}

Most existing methods for estimating causal effects rely on the ignorability assumption (no unobserved confounders), and adjust for all measurable confounders to derive the unbiased estimates of treatment effects. 
While Covariate Balancing Propensity Score (CBPS) (and its continuous extension)~\cite{imai2014covariateCBPS, fong2018covariateCBGPS} offers certain robustness without this assumption, it is limited to linear propensity model.
More common practices utilize proxy variables~\cite{cider2022, montgomery2000measuringproxies} \textendash\ measurable variables correlated to the unobserved confounders. While treating the proxies as ordinary confounders is incorrect and can lead to bias, other methods for unbiased estimation~\cite{pearl2012measurementpearlproxies} require strong assumptions about the structure of both the proxies and the hidden confounders (e.g., they are both categorical), which are often unrealistic in practical settings. 
Alternatively, Causal Effects Variational Auto-Encoder (CEVAE)~\cite{cevae17} proposed to model the hidden confounders as latent variables and use a Variational Auto-Encoder (VAE) to simultaneously discover the hidden confounders and how they affect treatment and outcome. By recovering the joint probability $p(\mathbf{X}, \mathbf{Z}, T, Y)$ of the hidden confounders, treatment, outcome and the observed covariates from observational data, the individual treatment effects become identifiable. However, the architecture of CEVAE is restricted to binary treatments, and few methods can address the challenges of unobserved confounders and continuous treatments simultaneously. 

%% file: 3_formulation.tex
\section{Problem Formulation}

\begin{wrapfigure}[18]{r}{0.3\textwidth}
\vspace{-1cm}
    \begin{center}
    \includegraphics[width=0.25\textwidth]{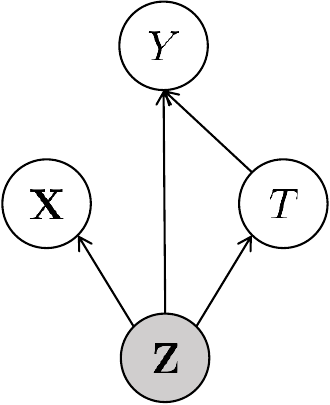}
    \end{center}
    
    \caption{\label{fig: causal_graph}The causal graph model. $T$ is a treatment, $Y$ is an outcome, $\mathbf{Z}$ is the unobserved confounders, and $\mathbf{X}$ is measurable covariates affected by $\mathbf{Z}$.}
\end{wrapfigure}

We assume a causal model of Figure~\ref{fig: causal_graph}. Unlike in~\cite{cevae17} where $T$ is a binary treatment, here $T$ takes continuous values, e.g., the dosage of a medication. $Y$ is an outcome, e.g., the recovery rate, $\mathbf{Z}$ is the unobserved confounders, e.g., living conditions, and covariates $\mathbf{X}$ are noisy proxies of $\mathbf{Z}$, e.g., blood pressures. Note that conditioned on $\mathbf{Z}$, $\mathbf{X}$ and $T$ is independent, so the joint probability can be factorized by $p(\mathbf{X}, \mathbf{Z}, T, Y) = p(\mathbf{Z}) p(\mathbf{X}|\mathbf{Z}) p(T|\mathbf{Z}) p(Y|\mathbf{Z}, T)$.

Suppose we receive observations $(\mathbf{x}_i, t_i, y_i)$ for $N$ samples, where $(\mathbf{x}_i, t_i, y_i)$ represents realisations of random variables $(\mathbf{X}, T, Y)$. Here, $\mathbf{x}_i \in \mathbb{R}^{d_x}$ is the covariate vector, $t_i$ is the observed treatment, and $y_i \in \mathbb{R}$ is the observed outcome of the $i$th sample. Without loss of generality, we assume the continuous treatment $T$ falls within the [0, 1] interval. The distribution of $(\mathbf{x}_i, t_i, y_i)$ is affected by unobserved confounders $\mathbf{z}_i$. Following the potential outcome framework~\cite{guo2020survey, rubin1984bayesianlypotentialoutcome}, we denote the potential outcome, which is the outcome $Y$ if applied an intervention $T=t$, as $Y|\text{Do}(T=t)$ or simply $Y(t)$. Unobserved potential outcomes are referred to as counterfactuals.

Our goal is to estimate the individual dose-response curve as
\begin{equation}
    \mu(\mathbf{x}, t) = \mathbb{E}[Y(t)|\mathbf{X}=\mathbf{x}],
\end{equation}
which is a direct extension of the individual treatment effect (ITE). ~\cite{cevae17} proved that with $p(\mathbf{X}, \mathbf{Z}, T, Y)$ recovered, the ITE of binary treatment $\mathbb{E}[Y(1)|\mathbf{X}=\mathbf{x}] - \mathbb{E}[Y(0)|\mathbf{X}=\mathbf{x}]$ is identifiable. Similarly, we prove in the Appendix~\ref{apdx: proof} that with the same premise, the individual dose-response curve $\mathbb{E}[Y(t)|\mathbf{X}=\mathbf{x}]$ is identifiable.

%% file: 4_method.tex
\section{Methodology}\label{sec: method}
We propose ContiVAE to estimate individual dose-response curve with the presence of hidden confounders by leveraging a variational auto-encoder (VAE) to learn the distribution of unobserved confounders as latent variables and approximately estimate $p(\mathbf{X}, \mathbf{Z}, T, Y)$. We first derive the variational lower bound of ContiVAE, then provide a detailed explanation of its neural network architecture and the design of the prior distribution. Finally, we briefly describe the training and inference process of ContiVAE.

\subsection{Deriving the Variational Lower Bound}
ContiVAE models the hidden confounders $\mathbf{Z}$ as latent variables, and utilizes VAE to estimate the relationships between $\mathbf{Z}$ and $(\mathbf{X}, T, Y)$. Specifically, we assume $\mathbf{Z}$ follows a prior distribution $p(\mathbf{Z})$. For each observed sample $(\mathbf{x}, t, y)$, the generative model takes a sample $\mathbf{z}$ to reproduce the observational data, maximizing

\begin{align}
    \log p(\mathbf{x}, t, y) &= 
    \log \int_{\mathbf{z}} p(\mathbf{x}, \mathbf{z}, t, y)\, d\mathbf{z} = \log \int_{\mathbf{z}} p(\mathbf{z})\, p(\mathbf{x}, t, y|\mathbf{z})\,d\mathbf{z}\nonumber\\
    &= \log \int_{\mathbf{z}} p(\mathbf{z}) \, p(\mathbf{x}|\mathbf{z}) \, p(t|\mathbf{z})\, p(y|t, \mathbf{z}) \, d\mathbf{z},
\label{eq:goal}
\end{align}
where the posteriors $p(\mathbf{x}|\mathbf{z})$, $p(t|\mathbf{z})$, $p(y|t, \mathbf{z})$ are parameterized by neural networks with parameters $\mathbf{\theta}$. As the prior $p(\mathbf{Z})$ is unknown, we approximate the posterior $p(\mathbf{z}|\mathbf{x}, t, y)$ with a neural network $q_\phi(\mathbf{z}|\mathbf{x}, t, y)$ which takes $(\mathbf{x}, t, y)$ as inputs and serves as the encoder of the VAE. Since minimizing the KL-divergence of $q_\phi(\mathbf{z}| \mathbf{x}, t, y)$ and $p(\mathbf{z}| \mathbf{x}, t, y)$ is to minimize
\begin{align}
    & \mathcal{D}\left[q_\phi(\mathbf{z}| \mathbf{x}, t, y) \parallel p(\mathbf{z}| \mathbf{x}, t, y) \right] \nonumber\\
    &= \mathbb{E}_{\mathbf{z} \sim q_\phi(\mathbf{z}| \mathbf{x}, t, y)}\left[ \log q_\phi(\mathbf{z}| \mathbf{x}, t, y) - \log p(\mathbf{z}| \mathbf{x}, t, y)  \right] \nonumber\\
    &= \mathbb{E}_{\mathbf{z} \sim q_\phi(\mathbf{z}| \mathbf{x}, t, y)} \left[ \log q_\phi(\mathbf{z}| \mathbf{x}, t, y) - \log p(\mathbf{z}) - \log p(\mathbf{x}, t, y | \mathbf{z}) + \log p(\mathbf{x}, t, y)\right] ,
\end{align}
the variational lower bound (ELBO) of our VAE model can be written as
\begin{align}
    \mathcal{L} = \sum_{i=1}^N \mathbb{E}_{\mathbf{z}_i \sim q_\phi(\mathbf{z}_i| \mathbf{x}_i, t_i, y_i)} \left[ \log p_\mathbf{\theta}(\mathbf{x}_i, t_i, y_i | \mathbf{z}_i)  
      + \log p(\mathbf{z}_i) - \log q_\phi(\mathbf{z}_i| \mathbf{x}_i, t_i, y_i) \right]]
\end{align}
\subsection{The Network Architecture}
With the ELBO derived, the next step is to specify neural networks $p_{\theta_x}(\mathbf{x}|\mathbf{z})$, $p_{\theta_t}(t|\mathbf{z})$, $p_{\theta_y}(y | t, \mathbf{z})$ and $q_\phi(\mathbf{z}|\mathbf{x}, t, y)$. Additionally, for an unseen sample with covariates $\mathbf{x}$, we need two auxiliary networks $q(t|\mathbf{x})$ and $q(y|t, \mathbf{x})$ to predict its treatment value and outcome respectively. 

\begin{figure}[t!]
    \centering
    \subfigure[Decoder.]{
        \includegraphics[width=.65\linewidth]{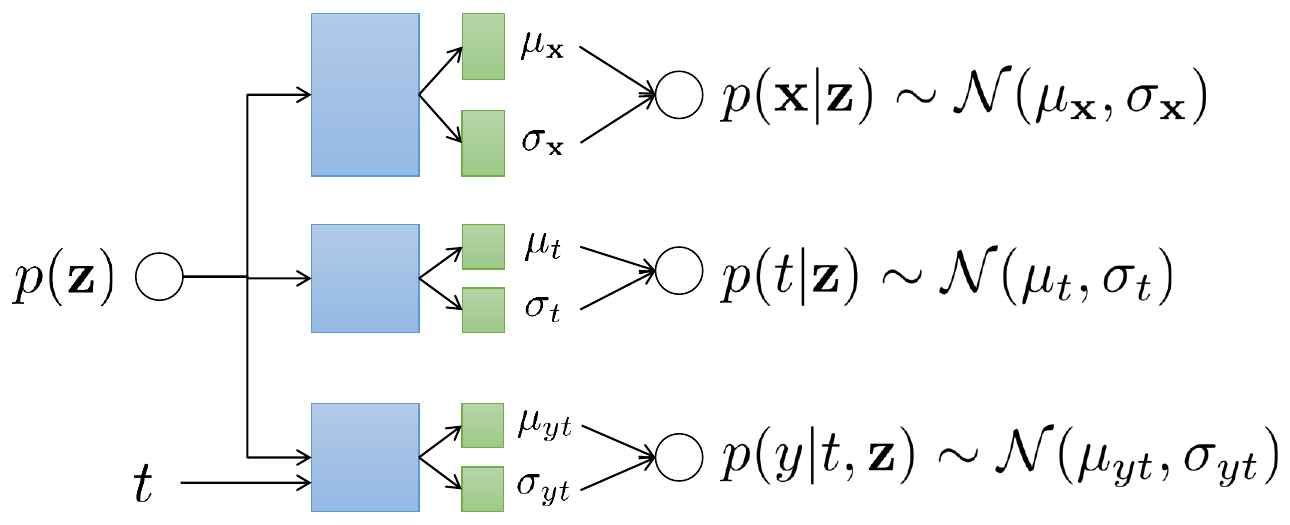}
        \label{fig: decoder}
    }
    \subfigure[Encoder.]{
        \includegraphics[width=.8\linewidth]{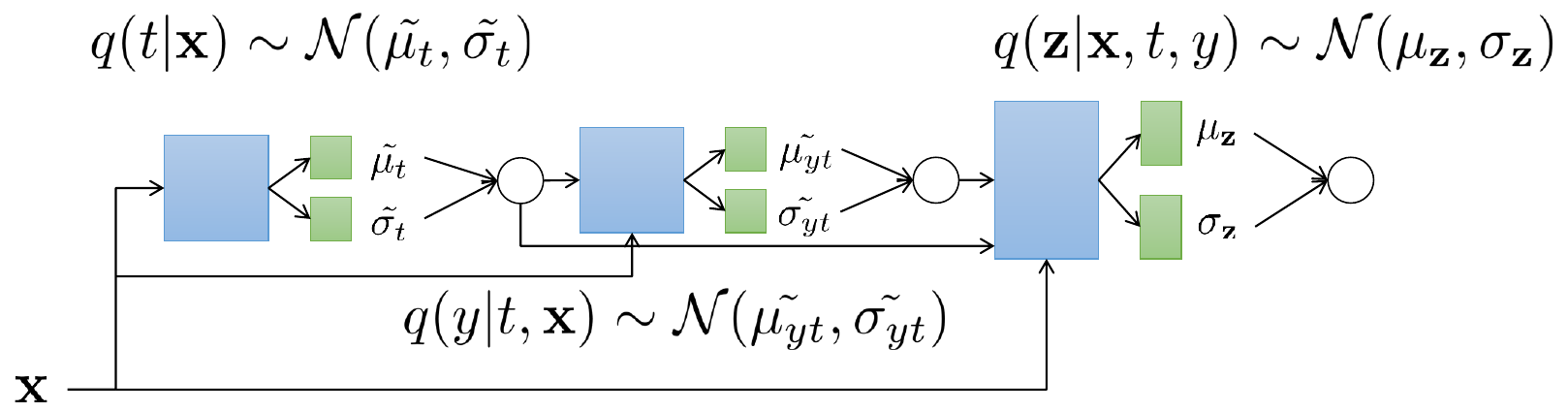}
        \label{fig: encoder}
    }
    \caption{The model architecture of ContiVAE. Blue rectangles represent MLP networks with $h$ units per hidden layer, while green rectangles denote output layers. Circles represent probability distributions.}
    \label{fig: model}
    \vspace{-0.2cm}
\end{figure}

The structure of ContiVAE is shown in Figure~\ref{fig: model}. Specifically, in the decoder(see Figure~\ref{fig: decoder}), we parameterize $p_{\theta_x}(\mathbf{x}|\mathbf{z})$ as a Gaussian distribution (or Bernoulli, etc., depending on the characteristics of each element in $\mathbf{X}$), with its mean and standard deviation estimated respectively by two output layers upon a shared multi-layer perceptions (MLP) network, taking $\mathbf{z}$ as the input:
\begin{equation}
    p_{\theta_x}(\mathbf{x}| \mathbf{z}) = \mathcal{N}(\mathbf{\mu}_\mathbf{x}, \mathbf{\sigma}_\mathbf{x}^2), \, \mathbf{\mu}_\mathbf{x} = f_1(\mathbf{z}), \, \mathbf{\sigma}_\mathbf{x} = \text{Softplus}(f_2(\mathbf{z})),
\end{equation}
The $\text{Softplus}$ activation ensures the resulting $\mathbf{\sigma}_\mathbf{x}$ is positive, $f_1$ and $f_2$ are neural networks. Similarly, as $T$ is a continuous variable, we model $p_{\theta_t}(t|\mathbf{z})$ as Gaussian, 
\begin{gather}
    p_{\theta_t}(t|\mathbf{z}) = \mathcal{N}(\mu_t, \sigma_t^2), \, \mu_t = f_3(\mathbf{z}), \, \sigma_t = \text{Softplus}(f_4(\mathbf{z})).
\end{gather}
To estimate the outcome $y$ corresponding to a given treatment value, the neural networks for $p_{\theta_y}(y|t, \mathbf{z})$ take the concatenated vector of $t$ and $\mathbf{z}$ as input:
\begin{gather}
    p_{\theta_y}(y|t, \mathbf{z}) = \mathcal{N}(\mu_{yt}, \sigma_{yt}^2), \, \mu_{yt} = f_5(t \parallel \mathbf{z}), \, \sigma_{yt} = \text{Softplus}\left(f_6(t \parallel \mathbf{z})\right).    
\end{gather}
where operation $t \parallel \mathbf{z}$ represents concatenating $t$ with $\mathbf{z}$. During training, the input $t$ for $p_{\theta_y}(y|t, \mathbf{z})$ is the observed treatment of each sample, while during inference, it corresponds to a treatment value of interest.

The structure of the encoder is represented in Figure~\ref{fig: encoder}. Following~\cite{cevae17}, we use $q(t|\mathbf{x})$ and $q(y|t, \mathbf{x})$ to predict $t$ and $y$ for an unseen sample $\mathbf{x}$:
{\small
\begin{gather}
    q(t|\mathbf{x}) = \mathcal{N}(\tilde{\mu_t}, \tilde{\sigma_t}^2), \, \tilde{\mu_t} = g_1(\mathbf{x}), \, \tilde{\sigma_t} = \text{Softplus}(g_2(\mathbf{x})).\\
    q(y|t,\mathbf{x}) = \mathcal{N}(\tilde{\mu_{yt}}, \tilde{\sigma_{yt}}^2), \, \tilde{\mu_{yt}} = g_3 \circ g_1(\mathbf{x}, \tilde t), \,  \tilde{\sigma_{yt}}= \text{Softplus}(g_4 \circ g_2(\mathbf{x}, \tilde t)).
\end{gather}
}%
Subsequently, $\mathbf{z}$ is encoded using the predicted results $\tilde t$ and $\tilde y$ along with $\mathbf{x}$:
{\small
\begin{equation}
        q(\mathbf{z}|\mathbf{x}, t, y) = \mathcal{N}(\mu_\mathbf{z}, \sigma_\mathbf{z}^2), \, \mu_\mathbf{z} = g_5 \circ g_3(\mathbf{x}, \tilde t, \tilde y), \, \sigma_\mathbf{z} = \text{Softplus}(g_6 \circ g_4(\mathbf{x}, \tilde t, \tilde y)).
\end{equation}
}%

The architecture of ContiVAE diverges significantly from CEVAE, which is inspired by TARNET~\cite{tarnet} and leverages distinct neural networks to model the probability distributions depending on $T$ under different treatment state ($t=1/0$). For instance, it adopts two neural networks trained on separate data to infer $p(y|1, \mathbf{z})$ and $p(y|0, \mathbf{z})$ respectively, taking solely $\mathbf{z}$ as input. Our design design contrasts with DRNets~\cite{schwab2020learningdrnet} as well, which extends TARNET by partitioning the dosage range of continuous treatments into multiple subintervals and adopts a similar multi-task strategy for each subinterval. Their design aims to prevent the diminishing influence of $t$ in high-dimensional hidden layers when directly inputted into a neural network. However, this strategy is unsuitable and unnecessary in our case. Since ContiVAE is trained to reconstruct $t$ in the decoder, the influence of $t$ and its information will be preserved in previous layers to ensure its reconstruction at the end. Therefore, we adopts single MLP networks for all treatment values, directly incorporating $t$ as part of the input. This design reserves flexibility and avoids introducing hyperparameters and additional complexity to model selection like in DRNets.

\subsection{Prior Distribution}
\begin{wrapfigure}[16]{r}{.5\textwidth}
\vspace{-1.5cm}
    \centering
    \includegraphics[width=0.9\linewidth]{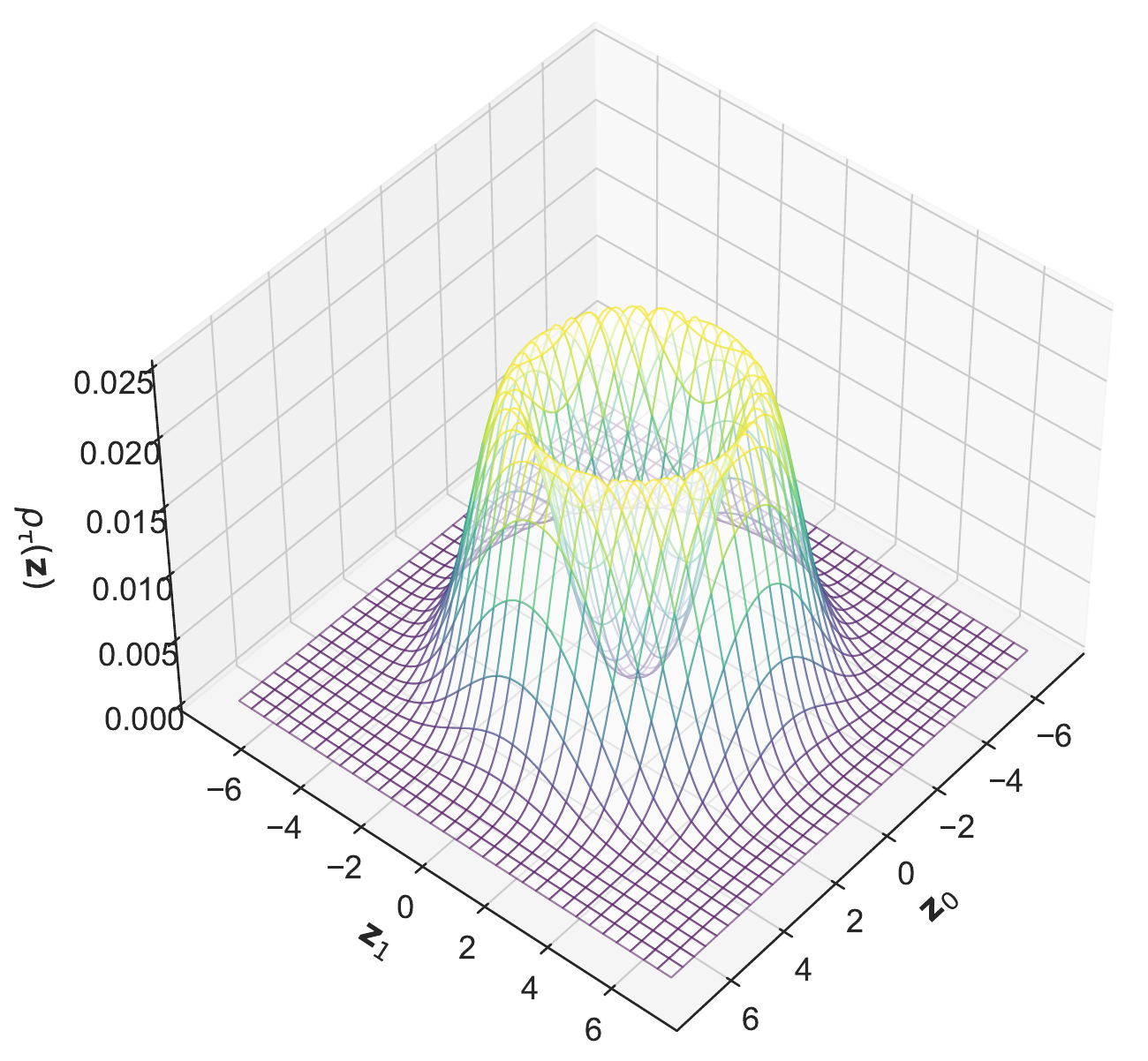}
    \caption{\label{fig:tilted}An example~\cite{floto2022tilted} of the Tilted Gaussian distribution in 2D, $\tau = 3$. The region of maximum density lies on the surface with a radius of 3.}
\end{wrapfigure}
Now we examine the prior distribution of $\mathbf{Z}$. 
ContiVAE adopts a Tilted Gaussian prior distribution~\cite{floto2022tilted}, rather than a standard Gaussian prior commonly employed in various methods~\cite{rezende2014stochasticvae, kingma2013autonormalvae} and in \cite{cevae17}.
Given our objective of estimating individual dose-response curves, which can vary significantly among individuals, the model must not only recover the curves of each sample 
but also capture individual heterogeneity at the same time. The normal distribution concentrates its high probability density region around a point, which becomes relatively smaller as the latent dimension increases. Minimizing the KL-divergence with a normal prior distribution will force the latent variables of different samples into the same point, thereby hindering the effective representation of individual heterogeneity. 
Opting for the Tilted Gaussian distribution helps address these issues.
The tilted Gaussian, denoted as $\mathcal{N}_\tau(0, \mathbf{I})$, is created by exponentially tilting the Gaussian distribution by its norm. In this distribution, the maximum probability density region lies on the surface of a sphere whose radius is $\tau$, rather than a single point. As the latent dimension increases, the volume of this region is exponentially larger compared to that of standard Gaussian as the latent dimension increases. The probability density function $\rho_\tau(\mathbf{z})$ of $\mathcal{N}_\tau(0, \mathbf{I})$ is defined as
\begin{equation}
    \rho_\tau(\mathbf{z}):= \frac{e^{\tau}\Vert\mathbf{z}\Vert}{\mathbf{Z}_\tau} \frac{e^{-\frac{1}{2}\Vert\mathbf{z}\Vert^2}}{\sqrt{2\pi} \mathbf{I}^{d_\mathbf{z}}}, \mathbf{Z}_\tau := \mathbb{E}_{\mathbf{z} \sim \mathcal{N}(0, \mathbf{I})}[e^{\tau \Vert \mathbf{z} \Vert}],
\end{equation}
where $d_\mathbf{z}$ is the dimension of $\mathbf{Z}$, and $\rho_0(\mathbf{z})$ is equal to the density function of $\mathcal{N}(0, \mathbf{I})$. An example of $\rho_\tau(\mathbf{z})$ in 2-D is visualized in Figure~\ref{fig:tilted}. More detailed information of the Tilted Gaussian distribution, readers are referred to~\cite{floto2022tilted}. With a tilted Gaussian prior, the KL-divergence term is calculated by
\begin{gather}
    \mathcal{D}\left[q_\phi(\mathbf{z}| \mathbf{x}, t, y) \parallel p(\mathbf{z}) \right] = \frac{1}{2}(\Vert \mu_\mathbf{z} \Vert - \Vert \mu_\tau^* \Vert)^2,\\
    \mu_\tau^* = \text{argmin}_\mu -\tau\sqrt{\frac{\pi}{2}}\mathbf{L}_{\frac{1}{2}}^{\frac{d_\mathbf{z}}{2}-1} \left(-\frac{\Vert \mu \Vert^2}{2} \right)+ \frac{\Vert \mu \Vert^2}{2},
\end{gather}
where $\mathbf{L}$ is the generalized Laguerre polynomial. Throughout the paper, we set $\tau$ to 3.

\subsection{Optimization Objective}
An investigation of CEVAE~\cite{rissanen2021criticalcevaefault} uncovered the problem of over-emphasizing the reconstruction of $\mathbf{X}$ when there exists high correlations in $\mathbf{X}$, which may lead to neglecting the hidden confounders and directly adjusting for covariates. We encountered the same problem when applying on real world datasets with very high-dimensional $\mathbf{X}$. To address this, we introduce a scaling parameter $\lambda \le 1.0$ to reduce the emphasis on the reconstruction loss of $\mathbf{X}$. The value of $\lambda$ will be selected through $k$-fold cross validation in practice. During training, we utilizes Stochastic Gradient Descent to minimize the loss function
\begin{align}
    \mathcal{L} &= \sum_{i=1}^N - (\,\lambda \cdot \log p_{\theta_x}(\mathbf{x}_i|\mathbf{z}_i) + \log p_{\theta_t}(t_i|\mathbf{z}_i) + \log p_{\theta_y}(y_i|t_i, \mathbf{z}_i) \nonumber\\ 
    &+ \log q(t_i|\mathbf{x}_i) + \log q(y_i|t_i, \mathbf{x}_i) \,) + \mathcal{D}\left[q_\phi(\mathbf{z}_i| \mathbf{x}_i, t_i, y_i) \parallel p(\mathbf{z}_i)\right].
\end{align}
During inference, to obtain individual dose-response curve of a given $\mathbf{x}$, we input it into the model and gather the expected output of $p(y|t, \mathbf{z})$ as the potential outcome at $t$.

%% file: 5_simulation_experiment.tex
\section{Experiments on Synthetic Datasets} \label{sec: simulation}
Due to the difficulty in obtaining real world datasets for causal inference with groundtruths, we evaluate our method on semi-synthetic datasets, seeking answers to the following questions: (1) How does the performance of ContiVAE compare to various baseline methods and state-of-the-art models? (2) How do varing hyperparameter configurations, especially choices of $\lambda$ influence ContiVAE's performance? (3) How does ContiVAE's performance change with increasing selection bias in the observational data?
\subsection{Experimental Setup}

\subsubsection{Datasets.} We conduct our experiments on two real world datasets: News~\cite{schwab2020learningdrnet} and The Cancer Genomic Atlas (TCGA)~\cite{weinstein2013cancertcga}\footnote{We use the same version of datasets as in \href{https://github.com/d909b/drnet}{https://github.com/d909b/drnet}.}. 

\paragraph{TCGA.} The TCGA dataset contains gene expression data of various types of tumors for 9,659 patients. Following ~\cite{scigan2020}, we select the 4,000 most variable genes for each patient as covariates $X$, and normalize them to the [0, 1] interval using min-max normalization. Each patient's covariate is further normalized to have norm 1. We consider the single treatment scenario and evaluate on 4 manually designed dose-response curves, as listed in Table~\ref{tab:tcga_curves}. Each dose-response curve $f^j(\mathbf{x}, t)$ is characterized by a set of parameters $\mathbf{v}_1^j$, $\mathbf{v}_2^j$, $\mathbf{v}_3^j$, which are randomly sampled from $\mathcal{N}(0, 1)$ and normalized to have norm 1. 

We randomly assign intervention $t$ to each patient, where $t$ is sampled from distribution $\text{Beta}(\alpha, \beta^j)$. Here, $\beta^j = \frac{\alpha-1}{t^{j*}} + 2 -\alpha$, where $t^{j*}$ is the optimal dosage for response curve $f^j$, ensuring that the mode of the beta distribution equals to $t^{j*}$. $\alpha$ determine the skewness of the distribution, thus controlling the dosage selection bias. When $\alpha=1$, $\text{Beta}(\alpha, \beta^j)$ reduces to a uniform distribution between 0 and 1. After assigning $t_i$ to a patient with covariate $\mathbf{x}_i$, its observed outcome is generated by adding a Gaussian noise $\epsilon \sim \mathcal{N}(0,0.02)$ to $f^j(\mathbf{x}_i, t_i)$. 

\begin{table}[t]
\begin{adjustbox}{center}
    \begin{tabular}{ccc}
    \Xhline{0.8pt}
    No. &  Dose-Response Curve & Optimal Dosage \\
    \specialrule{0.7pt}{1pt}{1pt}
    \multirow{2}*{1}  &  \scalebox{0.9}{\multirow{2}*{$ C\left((\mathbf{v}_1^1)^T \mathbf{x} + 12t\left(t- k\right)^2\right), \text{where } k = 0.75 \frac{(\mathbf{v}_2^1)^T \mathbf{x}}{(\mathbf{v}_3^1)^T 
    \mathbf{x}}$}} & \scalebox{0.9}{$\frac{k}{3}$ if $k\ge0.75$}\\
    & & \scalebox{0.9}{1 if $k < 0.75$} \\
    
    \specialrule{0.5pt}{1pt}{1pt}
    2 & \scalebox{0.9}{$C\left( (\mathbf{v}_1^2)^T \mathbf{x} + \sin \left(\pi \left(\frac{\mathbf{v}_2^{2T}\mathbf{x}}{\mathbf{v}_3^{2T}\mathbf{x}}\right) t\right)\right)$} & \scalebox{0.9}{$t_2^* =\frac{(\mathbf{v}_3^{2})^T \mathbf{x}}{2 (\mathbf{v}_2^{2})^T \mathbf{x}}$}\\
    \specialrule{0.5pt}{1pt}{1pt}
    3 & \scalebox{0.9}{$ C \left( (\mathbf{v}_1^3) ^T \mathbf{x} + 12 (\mathbf{v}_2^3) ^T \mathbf{x}t - 12 (\mathbf{v}_3^3)^T \mathbf{x} t^2 \right)$} & \scalebox{0.9}{$t_3^*=\frac{(\mathbf{v}_2^3)^T \mathbf{x}}{ 2 (\mathbf{v}_3^3)^T \mathbf{x}}$ }\\
    
    \specialrule{0.5pt}{1pt}{1pt}
    \multirow{3}*{4} & \multirow{3}*{\scalebox{0.9}{$ C \left( \cos ( (2+ (\mathbf{v}_1^4)^T \mathbf{x} ) \pi t  + (\mathbf{v}_2^4)^T \mathbf{x} \pi ) + (\mathbf{v}_3^4)^T \mathbf{x}\right) $}} & \scalebox{0.9}{$b$ if $b = \frac{2-\mathbf{v}_2^{4T} \mathbf{x}}{ 2 + \mathbf{v}_1^{4T}\mathbf{x}} \le 1$}\\
    & & \scalebox{0.9}{0 if $\cos (\pi (\mathbf{v}_2^4) ^T \mathbf{x}) > \cos(\pi (2+ (\mathbf{v}_1^4) ^T\mathbf{x} + (\mathbf{v}_2^4)^T\mathbf{x}))$} \\
    & & \scalebox{0.9}{1 else.}\\
    \specialrule{0.8pt}{1pt}{1pt}
    \end{tabular}
    \end{adjustbox}
    \caption{The 4 dose-response curves used to generate outcomes using covariates $\mathbf{x}$ from TCGA dataset, and their corresponding optimal dosage $t^*$. When $t = t^{j*}$, the individualized dose-response curve $f^j(\mathbf{x}, t)$ reaches its maximum value.}
    \label{tab:tcga_curves}
    \vspace{-0.3cm}
\end{table}

\begin{table}[t]
    \begin{adjustbox}{center}
    \begin{tabular}{ccc}
    \Xhline{0.8pt}
    No. &  Dose-Response Curve & Optimal Dosage \\
    \specialrule{0.7pt}{1pt}{1pt}
    \multirow{2}*{1}  & \scalebox{0.9}{ \multirow{2}*{$C\left( \mathbf{u}^T \mathbf{V}_1^1 \mathbf{x} + 12t\left(t- k\right)^2\right), \text{where } k = 0.75 \frac{ \mathbf{u}^T \mathbf{V}_2^1\mathbf{x}}{\mathbf{u}^T \mathbf{V}_3^1 
    \mathbf{x}}$}} & \scalebox{0.9}{$\frac{k}{3}$ if $k\ge0.75$}\\
    & & \scalebox{0.9}{1 if $k < 0.75$} \\
    
    \specialrule{0.5pt}{1pt}{1pt}
    2 & \scalebox{0.9}{$ C\left( \mathbf{u}^T \mathbf{V}_1^2 \mathbf{x} + \sin \left(\pi \left(\frac{\mathbf{u} ^T \mathbf{V}_2^2 \mathbf{x}}{\mathbf{u}^T \mathbf{V}_3^2\mathbf{x}}\right) t\right)\right)$} & \scalebox{0.9}{$t_2^* =\frac{\mathbf{u}^T \mathbf{V}_3^2 \mathbf{x}}{2 \mathbf{u}^T \mathbf{V}_2^2 \mathbf{x}}$}\\
    
    \specialrule{0.5pt}{1pt}{1pt}
    3 & \scalebox{0.9}{$C \left( \mathbf{u}^T \mathbf{V}_1^3 \mathbf{x} + 12 \mathbf{u}^T \mathbf{V}_2^3 \mathbf{x}t - 12 \mathbf{u}^T \mathbf{V}_3^3 \mathbf{x} t^2 \right)$} & \scalebox{0.9}{$t_3^*=\frac{\mathbf{u}^T \mathbf{V}_2^3 \mathbf{x}}{ 2 \mathbf{u}^T \mathbf{V}_3^3 \mathbf{x}}$} \\
    
    \specialrule{0.5pt}{1pt}{1pt}
    \multirow{3}*{4} & \scalebox{0.9}{\multirow{3}*{$C \left( \cos ( (2+ \mathbf{u}^T \mathbf{V}_1^4 \mathbf{x} ) \pi t  + \mathbf{u}^T \mathbf{V}_2^4 \mathbf{x} \pi ) + \mathbf{u}^T \mathbf{V}_3^4 \mathbf{x}\right) $}}& \scalebox{0.9}{$b$ if $b = \frac{2-\mathbf{u}^T \mathbf{V}_2^4 \mathbf{x}}{ 2 + \mathbf{u}^T \mathbf{V}_1^4 \mathbf{x}} \le 1$} \\
    & & \scalebox{0.9}{0 if $\cos (\pi \mathbf{u}^T \mathbf{V}_2^4 \mathbf{x}) > \cos(\pi (2+ \mathbf{u}^T \mathbf{V}_1^4 \mathbf{x} + \mathbf{u}^T \mathbf{V}_2^4 \mathbf{x}))$} \\
    & & \scalebox{0.9}{1 else.} \\
    \specialrule{0.8pt}{1pt}{1pt}
    \end{tabular}
    \end{adjustbox}
    \caption{The 4 dose-response curves used to generate outcomes using covariates $\mathbf{x}$ and hidden confounders $\mathbf{u}$ from News dataset, and their corresponding optimal dosage $t^*$. When $t = t^{j*}$, the individualized dose-response curve $f^j(\mathbf{x}, \mathbf{u}, t)$ reaches its maximum value.}
    \label{tab:news_curves}
    \vspace{-0.5cm}
\end{table}

\paragraph{News.} The News dataset comprises 100,000 news articles sampled from the NY Times corpus. Each news article is associated with a 2,876-dimensional bag-of-word vector, serving as the covariates $\mathbf{X}$, as well as a 50-dimensional vector which can be think of as the browsing information of the reader, considered as the hidden confounders $\mathbf{U}$.  We adopt the same method as with the TCGA dataset to generate observed data, and also examine four dose-response curves. However, while the dose-response curves on the TCGA dataset are functions of $\mathbf{x}$ and $t$, we explicitly set the dose-response curves on News dataset to be dependent on $\mathbf{u}$ by modifying paramters $\mathbf{v}_1^j, \mathbf{v}_2^j, \mathbf{v}_3^j$ to $\mathbf{u}^T \mathbf{V}_1^j, \mathbf{u}^T \mathbf{V}_2^j, \mathbf{u}^T \mathbf{V}_3^j$, where $\mathbf{V}_1^j, \mathbf{V}_2^j, \mathbf{V}_3^j$ are matrices sampled from $\mathcal{N}^{50 \times 2876}$. The definition of the 4 dose-response curves are provided in Table~\ref{tab:news_curves}. Note that the hidden confounders $\mathbf{U}$ is not visible to models during training; they are solely used for data generation and obtaining groundtruths.

\subsubsection{Models.}

\begin{itemize}
\item \textbf{ContiVAE and ContiVAE-N.} In addition to our proposed ContiVAE, to investigate the performance enhancement of using Tilted-Gaussian distribution as the prior latent distribution, we experiment with a variant of ContiVAE that utilizes a normal latent distribution, referred to as ContiVAE-N.
\item \textbf{MLP.} As a baseline, we develop an multi-layer perceptron (MLP) network which takes covariates and dosages as input and predicts the outcomes.
\item \textbf{SCIGAN.} SCIGAN~\cite{scigan2020} uses GANs to estimate individualized dose-response curves for multiple continuous treatments. It trains a counterfactual generator and discriminator adversarially, then utilizes the trained generator to supervisely train an inference network, achieving the state-of-art performance. In this work, we focus on single treatment scenario, so we set the number of treatments in SCIGAN to 1 and use optimal hyperparameters given in~\cite{scigan2020}.
    \item \textbf{GPS.} Generalized Propensity Score (GPS)~\cite{imai2004causalgps} extends Propensity Score Matching to continuous treatment settings. We implement GPS using the \verb|causal_curve| python package with a normal treatment model and 2nd degree polynomial outcome model.
\item \textbf{CBGPS.} Covariate Balancing Generalized Propensity Score (CBGPS)~\cite{fong2018covariateCBGPS} is an improvement of GPS that estimating the propensity score by optimizing the resulting covariate balancing. It increases robustness against model misspecification, and improves covariate balancing even in the presence of hidden confounders. 
We implement CBGPS using the \verb|CBPS| R package, with a linear treatment model and 2nd degree polynomial regression.
    
\end{itemize}

\subsubsection{Hyperparameters.} We select the best hyperparameters of MLP and our model for each dataset and dose-response curve through 5-fold cross validation, and use the best settings given in the original paper~\cite{scigan2020} for SCIGAN. 
On News dataset, we train all models using Adam~\cite{kingma2014adam} optimizer at a learning rate of 0.0001 for 100 epochs, and on TCGA dataset with learning rate of 0.001 for 300 epochs. 

\subsubsection{Metrics.}
For each dataset and dose-response curve, we compare the models by the root of Mean Integrated Square Error (MISE) and Dosage Policy Error (DPE). MISE, defined in Eq.~\ref{eq: mise}, measures the overall accuracy of the model in estimating potential outcomes across the entire dosage range: 
\begin{equation}\label{eq: mise}
    \text{MISE} = \frac{1}{N} \sum_{i=1}^{N} \int \left(y_i(t) - \hat y_i(t) \right)^2 dt
\end{equation}
DPE evaluates how accurately the model predicts the optimal dosage for each patient:
\begin{equation}\label{eq: dpe}
    \text{DPE} = \frac{1}{N}\sum_{i=1}^{N} \left( y_i(t^*) -y_i(\hat t^*)\right)^2,
\end{equation}
where $t^*$ is the true optimal dosage, $\hat t^*$ is the predicted optimal dosage. 
All the results are obtained on a held-out test set comprising 20\% of the data, and averaged over 5 repeat runs.


\begin{table}[h!]
    \centering
    \begin{tabular}{c|ccccc}
    \specialrule{1.3pt}{0pt}{0.5pt}
       Method & & Curve 1 & Curve 2 & Curve 3 & Curve 4 \\
    \specialrule{0.8pt}{0.5pt}{0.5pt}
       \multirow{2}*{ContiVAE} & \scalebox{0.8}{$\sqrt{\text{MISE}}$} 
                           &  \textbf{0.2593 $\pm$ 0.06} 
                           &  \textbf{0.1401 $\pm$ 0.01} 
                           &  $0.3718 \pm0.12$
                           &  \textbf{0.3035$\pm$0.06} \\
                           & \scalebox{0.8}{$\sqrt{\text{DPE}}$}
                           & \textbf{0.3204$\pm$0.07}
                           & $0.0043\pm0.002$
                           & $0.0084 \pm0.004$
                           & \textbf{0.0197 $\pm$ 0.003} \\
        \multirow{2}*{ContiVAE-N} & \scalebox{0.8}{$\sqrt{\text{MISE}}$} 
                                  & \textit{0.4471 $\pm$ 0.08}
                                  & $0.2915 \pm 0.07$
                                  & $0.6348 \pm 0.05$
                                  & \textit{0.8165 $\pm$ 0.06} \\
                                  & \scalebox{0.8}{$\sqrt{\text{DPE}}$}
                                  & \textit{0.2171 $\pm$ 0.60}
                                  & $0.0060 \pm 0.01$
                                  & $0.0143 \pm 0.02$
                                  & \textit{0.0603 $\pm$ 0.04} \\
    \specialrule{0.8pt}{0.5pt}{0.5pt}
        \multirow{2}*{SCIGAN} & \scalebox{0.8}{$\sqrt{\text{MISE}}$}
                              & 1.5720 $\pm$ 0.12
                              & $0.9367 \pm 0.80$
                              & $1.2003 \pm0.46$
                              & 3.7088 $\pm$ 1.07 \\
                              & \scalebox{0.8}{$\sqrt{\text{DPE}}$}
                              & $1.7665\pm 2.34$
                              & $0.0108\pm 0.01$
                              & $0.0102 \pm 0.01$
                              & $1.9687 \pm1.00$ \\
    \specialrule{0.8pt}{0.5pt}{0.5pt}
        \multirow{2}*{MLP} & \scalebox{0.8}{$\sqrt{\text{MISE}}$}
                           & $1.6674 \pm 0.66$
                           & $3.6899 \pm 0.18$
                           & $8.0742\pm 0.27$
                           & $7.5790\pm 3.07$ \\
                           & \scalebox{0.8}{$\sqrt{\text{DPE}}$}
                           & 0.4805 $\pm$ 0.40
                           & $9.9710 \pm 0.06$
                           & $22.4423\pm 0.00$
                           & $13.9378 \pm 8.73$ \\
    \specialrule{0.8pt}{0.5pt}{0.5pt}
        \multirow{2}*{GPS} & \scalebox{0.8}{$\sqrt{\text{MISE}}$}
                           & $3.6592 \pm 0.43$
                           & $0.3634 \pm 0.02$
                           & \textit{0.3716 $\pm$ 0.04}
                           & $8.1514 \pm 0.30$ \\
                           & \scalebox{0.8}{$\sqrt{\text{DPE}}$}
                           & $7.3813 \pm 0.24$
                           & \textit{0.0010 $\pm$ 0.001}
                           & \textit{0.0026 $\pm$ 0.003}
                           & $0.1723 \pm 0.03$ \\
    \specialrule{0.8pt}{0.5pt}{0.5pt}
        \multirow{2}*{CBGPS} & \scalebox{0.8}{$\sqrt{\text{MISE}}$}
                             & $2.7198 \pm 0.33 $
                             & \textit{0.2354 $\pm$ 0.01 }
                             & \textbf{0.2500 $\pm$ 0.03 }
                             & $6.8300 \pm 0.27 $ \\
                             & \scalebox{0.8}{$\sqrt{\text{DPE}}$}
                             & $6.3100 \pm 3.22 $
                             & \textbf{0.0007 $\pm$ 0.00}
                             & \textbf{0.0023 $\pm$ 0.002}
                             & 0.1521 $\pm$ 0.07 \\
    \specialrule{1.3pt}{0.5pt}{2pt}
    \end{tabular}
    \caption{Performances of all compared models for four curves on TCGA dataset. Metrics are reported as mean $\pm$ half-width of 0.95 confidence interval. The best results are highlighted in bold while the second best are italicized.}
    \label{tab:tcga_results}
    \vspace{-0.3cm}
\end{table}

\begin{table}[t!]
    \centering
    \begin{tabular}{c|ccccc}
    \specialrule{1.3pt}{0pt}{0.5pt}
       Method & & Curve 1 & Curve 2 & Curve 3 & Curve 4 \\
    \specialrule{0.8pt}{0.5pt}{0.5pt}
       \multirow{2}*{ContiVAE} & \scalebox{0.8}{$\sqrt{\text{MISE}}$}
                           & \textbf{1.2357$\pm$ 0.09}
                           & \textbf{0.7534 $\pm$ 0.03}
                           & \textbf{0.0885 $\pm$0.004}
                           & \textbf{0.7849$\pm$0.02} \\
                           & \scalebox{0.8}{$\sqrt{\text{DPE}}$}
                           & \textit{2.8568 $\pm$0.34}
                           & \textbf{0.1156 $\pm$0.01}
                           & \textit{0.0038$\pm$0.001}
                           & \textbf{0.2856$\pm$ 0.09} \\
        \multirow{2}*{ContiVAE-N} & \scalebox{0.8}{$\sqrt{\text{MISE}}$}
                                  & $1.5663 \pm 0.19$
                                  & $0.7868 \pm 0.02$
                                  & $0.0958 \pm 0.01$
                                  & $1.0697 \pm 0.05$ \\
                                  & \scalebox{0.8}{$\sqrt{\text{DPE}}$}
                                  & $3.1903 \pm 0.57 $
                                  & $0.1364 \pm 0.004$
                                  & $0.0041 \pm 0.001$
                                  & \textit{0.2993 $\pm$ 0.09} \\
    \specialrule{0.8pt}{.5pt}{.5pt}
        \multirow{2}*{SCIGAN} & \scalebox{0.8}{$\sqrt{\text{MISE}}$}
                              & $2.8740 \pm 0.30$
                              & $1.6702 \pm 0.56$
                              & $0.1740 \pm 0.05$
                              & $3.4695 \pm 0.45$ \\
                              & \scalebox{0.8}{$\sqrt{\text{DPE}}$}
                              & $3.5969 \pm 2.84$
                              & 0.3307 $\pm$ 0.28
                              & $0.0252  \pm 0.03$
                              & $0.8911 \pm0.56$ \\
    \specialrule{0.8pt}{.5pt}{.5pt}
        \multirow{2}*{MLP} & \scalebox{0.8}{$\sqrt{\text{MISE}}$}
                           & \textit{1.2619 $\pm$ 0.04}
                           & \textit{0.7716 $\pm$ 0.01}
                           & $0.1244 \pm 0.003$
                           & \textit{0.8615 $\pm$ 0.01} \\
                           & \scalebox{0.8}{$\sqrt{\text{DPE}}$}
                           & \textbf{2.7795 $\pm$ 0.40}
                           & \textit{0.1279 $\pm$ 0.005}
                           & $0.0275 \pm 0.002$
                           & 0.3631 $\pm$ 0.08 \\
    \specialrule{0.8pt}{.5pt}{.5pt}
        \multirow{2}*{GPS} & \scalebox{0.8}{$\sqrt{\text{MISE}}$}
                           & $3.4418 \pm 1.15$
                           & $0.8472 \pm 0.04$
                           & \textit{0.0907 $\pm$ 0.001 }
                           & $6.1522 \pm 0.04$ \\
                           & \scalebox{0.8}{$\sqrt{\text{DPE}}$}
                           & $8.3123 \pm 0.75$
                           & 0.2361 $\pm$ 0.03
                           & \textit{0.0038 $\pm$ 0.002}
                           & $8.6744 \pm 0.05$ \\
    \specialrule{.8pt}{.5pt}{.5pt}
        \multirow{2}*{CBGPS} & \scalebox{0.8}{$\sqrt{\text{MISE}}$}
                             & $2.9787 \pm 0.22 $
                             & $0.8325 \pm 0.08 $
                             & $0.0921 \pm 0.001$
                             & $6.1483 \pm 0.03 $ \\
                             & \scalebox{0.8}{$\sqrt{\text{DPE}}$}
                             & $7.7260 \pm 0.07 $
                             & $0.2567 \pm 0.31 $
                             & \textbf{0.0035 $\pm$ 0.001}
                             & $8.6988 \pm 0.02 $ \\
    \specialrule{1.3pt}{.5pt}{2pt}
    \end{tabular}
    \caption{Performances of all compared models for four curves on News dataset.}
    \label{tab:news_results}
    \vspace{-0.6cm}
\end{table}

\subsection{Benchmark Comparison}
The experimental results of all compared models on the TCGA dataset and the News dataset are represented in Table~\ref{tab:tcga_results} and~\ref{tab:news_results} respectively. We found that ContiVAE consistently achieves the best or comparable performance across both datasets and all evaluated curves, with significantly lower $\sqrt{\text{MISE}}$ and $\sqrt{\text{DPE}}$ than other methods. This demonstrates ContiVAE's robustness and its ability to accurately estimate individual dose-response curves. Notably, ContiVAE with a Tilted Gaussian prior exhibits significant improvement over its variant with the Normal prior (ContiVAE-N), indicating that changing to Tilted Gaussian expands the high probability density region in latent space, enabling the model to better capture individual differences and handle complex curve shapes. 
    
We also observed that parametric methods like GPS and CBGPS perform well then the underlying curve aligns with their assumptions (Curve 3). However, for complex curves (Curve 1 and 4), neural networks, including ContiVAE, significantly outperform them. 
ContiVAE's strengths also lies in its flexibility, 
achieving competitive results in both scenarios.

While MLP performs comparably to ContiVAE on the large News dataset (100,000), but its performance suffers significantly on the smaller TCGA dataset (9,659). This suggests that MLP requires more training data to converge while ContiVAE is more data-efficient. The results of MLP and ContiVAE on the News dataset with reduced sample size (see Appendix~\ref{apdx: mlpresults}) illustrate MLP's dependence on larger sample size.

\subsection{Parameter Sensitivities}
\begin{figure}[thbp]
    \centering
    \includegraphics[width=1.0\linewidth]{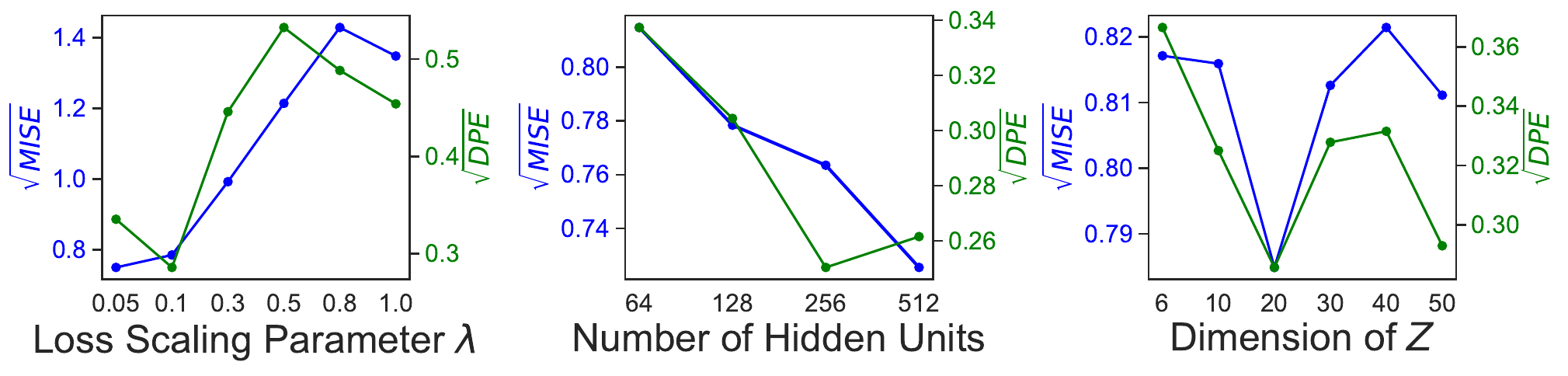}
    \caption{Performance of our model on Curve 4, News dataset with different settings of $\lambda$ in the loss function, the number of units in each hidden layer and the dimension of latent variable $Z$. The blue line and y-axis on the left represents $\sqrt{\text{MISE}}$, and the green line and y-axis on the right represents $\sqrt{\text{DPE}}$.}
    \label{fig: parameter}
\end{figure}

We evaluate ContiVAE's sensitivity to key parameters, especially the scaling parameter $\lambda$, using 5-fold cross validation on Curve 4 of the News dataset (Figure~\ref{fig: parameter}). 
Notably, introducing $\lambda$ resolves the previously identified issue~\cite{rissanen2021criticalcevaefault} and decreasing $\lambda$ significantly enhances ContiVAE's performance. Increasing hidden units 
generally improves performance, with $h=128$ being sufficient. 
Lastly, while a high-dimensional latent space is crucial for real-world data complexity, excessively high dimensions can hinder correct estimation. In our case, $d_\mathbf{z}=20$ yield optimal results.
\vspace{-1cm}
\subsection{Robustness to Selection Bias}
\begin{figure}[h!]
\vspace{-0.5cm}
    \centering
    \includegraphics[width=\linewidth]{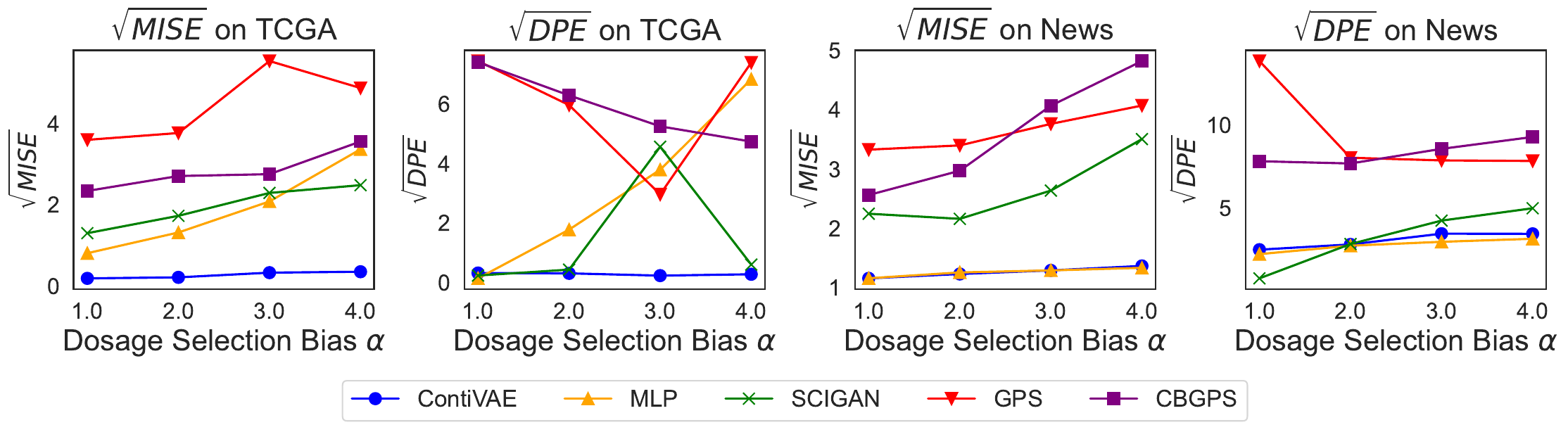}
    \caption{Performance of all models with increasing dosage selection bias $\alpha$ on Curve 1 of TCGA dataset and News dataset. When $\alpha = 1.0$, dosage assignment is a uniform distribution in [0, 1]. When $\alpha$ grows, the difference between $\beta$ and $\alpha$ increases, thus making the dosage distribution $\text{Beta}(\alpha, \beta)$ more skewed.}
    \label{fig:selection_all}
    \vspace{-0.3cm}
\end{figure}
To assess the robustness of each model to increasing levels of dosage selection bias, we evaluate their performance on Curve 1 of both datasets with varing $\alpha$ from 1.0 to 4.0 (Figure~\ref{fig:selection_all}. 
When there is no selection bias ($\alpha$=1), all neural network-based methods (ContiVAE, MLP and SCIGAN) yield similar results. However, ContiVAE distinguishes itself by maintaining consistent performance across all $\alpha$ values, significantly outperforming other models as the bias increased.

%% file: 6_volunteer.tex
\section{Application on Volunteer Data}\label{sec:volunteer}
In this section, we apply ContiVAE to estimate the effects of a volunteer's social network position on retention. Here, the social network is formed as volunteers engage in various tasks together, with their position quantified by social network metrics like degree centralities. Volunteer retention indicates how actively a volunteer will continue engaging in voluntary activities in the future, and can be quantified by the number of tasks a volunteer undertakes in an upcoming time span. The retention of volunteers is a crucial issue faced by non-profit and community organizations~\cite{volunteer2003, al2015volunteerimpact, cuskelly2006volunteermanagement}. 
Applying ContiVAE, we aim to answer to which extent, the different position a volunteer occupies in the social network would affect her future retention.

\subsection{Volunteer Data}
Our volunteer data is collected from Anti-Pandemic Pioneer, a mobile crowdsourcing platform launched during the COVID-19 epidemic to organize volunteer activities in Shenzhen, China. We collected 708,933 task participation logs from February 2020 to February 2022, involving 160,677 volunteers, each containing participation details such as timestamps, locations, task descriptions, volunteer IDs, and task IDs. Volunteer profiles, including age and task preferences, etc., are also collected as covariates $\mathbf{X}$. 
We split the time span into two segments. In the first segment, we construct a network $\mathcal{G}(\mathcal{V}, \mathcal{E}, \mathcal{W})$ of volunteers based on their co-participation relationships, where $v_i$ represents a volunteer, the presence of an edge $e_{ij}$ indicates that $v_i$ and $v_j$ collaborated at least once, and the edge weight $w_{ij}$ denotes the frequency of co-participation. The treatment $T$ of a volunteer is one of the indicators including centrality and structural hole spanner metrics~\cite{burt}, 
calculated from the constructed graph. For each volunteer, we quantify the observed outcome $Y$ by counting the number of her participated tasks in the second segment. 

\begin{figure}[t!]
    \centering
    \subfigure[]{
        \includegraphics[width=0.3\linewidth]{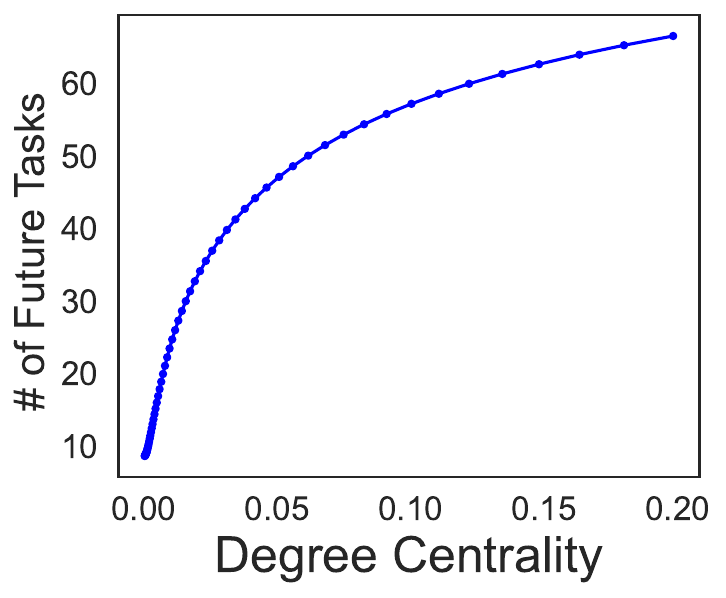}
    }
    \subfigure[]{
        \includegraphics[width=0.3\linewidth]{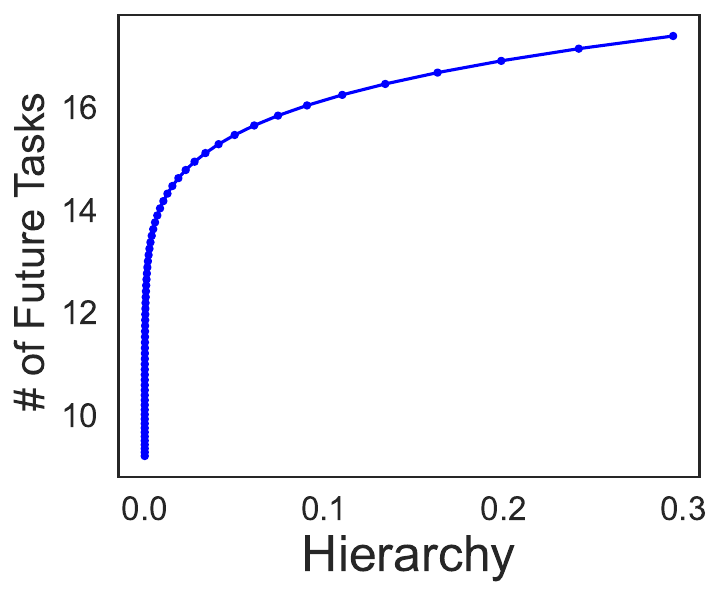}
    }
    \subfigure[]{
        \includegraphics[width=0.297\linewidth]{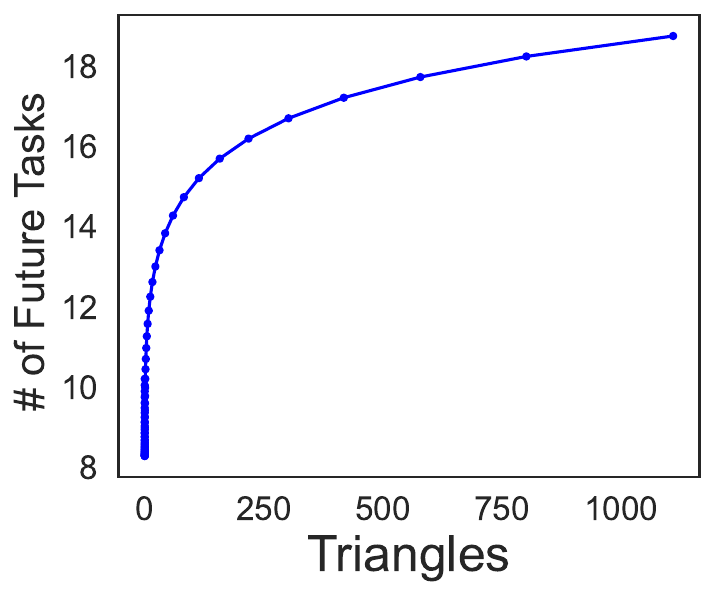}
    }
    \vspace{-0.3cm}
    \caption{Averaged dose-response curves of the number of a volunteer's future participation with (a) the degree centrality, (b) the hierarchy, (c) the number of triangles in the ego-network.}
    \label{fig:volunteer_result}
    \vspace{-0.3cm}
\end{figure}
\subsection{Results}
From the dose-response curves predicted by ContiVAE, we find that volunteers with higher degree centrality, hierarchy or number of triangles in ego-network exhibit a higher intention to remain engaged. We present the averaged curves depicting how the number of future tasks varies with these indicators across all volunteers in Figure~\ref{fig:volunteer_result}. 
The degree centrality of a volunteer is the number of collaborators normalized by the total number of volunteers. Hierarchy, as defined by Burt~\cite{burt}, is an indicator to measure the quality of structural hole spanners \textendash \ individuals that bridge otherwise disconnected groups in a social network, such as scientists collaborating across disciplines to connect researchers from diverse fields. A higher hierarchy value indicates a volunteer's collaborations are more concentrated on a single collaborator. Finally, the number of triangles in a volunteer's ego-network captures the density of connections among their collaborators. As Figure~\ref{fig:volunteer_result} suggests, occupying central position through collaboration with numerous colleagues strongly encourages a volunteer to continue participation. In addition, fostering a strong tie with a single collaborator or building a tightly-knit group also promotes volunteer engagement.

%% file: 00_appendix.tex
\begin{subappendices}
\renewcommand{\thesection}{\Alph{section}}
\section{Identifiability of Individual Dose-Response Curves}\label{apdx: proof}
\begin{theorem}
If $p(\mathbf{X}, \mathbf{Z}, T, Y)$ is recovered, $\mathbb{E}[Y(t)|\mathbf{X} = \mathbf{x}]$ is identifiable from observational data under the causal model in Figure~\ref{fig: causal_graph}.
\end{theorem}
\begin{proof}
    We prove that $p(Y|\mathbf{X}, \text{do}(t))$ is identifiable if $p(\mathbf{X}, \mathbf{Z}, T, Y)$ is recovered, then $\mathbb{E}[Y(t)|\mathbf{X}] = \int_y  y \cdot p(y| \mathbf{X}, \text{do}(t))\, dy$ is identifiable.
    \begin{align}
        p(Y|\mathbf{X}, \text{do}(t)) & \overset{1}= \int_\mathbf{Z} p(y|\mathbf{X}, \text{do}(t), \mathbf{Z}) \,p(\mathbf{Z}|\mathbf{X}, \text{do}(t)) \, d\mathbf{Z}\\
        & \overset{2}= \int_\mathbf{Z} p(y|\mathbf{X}, t, \mathbf{Z}) \, p(\mathbf{Z}|\mathbf{X}) \, d\mathbf{Z}\label{eq: terms}
    \end{align}
    $\overset{1}=$ is given by Pearl's backdoor adjustment formula~\cite{pearl2016causalprimer}, and $\overset{2}=$ is by applying do-operation on $T$ in the causal graph, which makes $Z$ independent of $T$ and remains the relationship of $Y$ and other variables. Once $p(\mathbf{X}, \mathbf{Z}, T, Y)$ is recovered, expression~\ref{eq: terms} can be calculated, thereby completing our proof.
\end{proof}

\section{Experimental Results of MLP and ContiVAE on News Dataset with Small Sample Size}\label{apdx: mlpresults}
We maintain the same data simulation parameters and the test set size of the News dataset, but reduce the size of training set (0.8 $\times$ 100,000 samples) to match that of the TCGA dataset (0.8 $\times$ 9,695 samples). We then perform experiments using ContiVAE and MLP, and the results are listed in Table~\ref{tab:appendixmlp}. 

\begin{table}[h!]
    \centering
    \begin{tabular}{c|ccccc}
    \specialrule{1.3pt}{0pt}{0.5pt}
       Method & & Curve 1 & Curve 2 & Curve 3 & Curve 4 \\
    \specialrule{0.8pt}{0.5pt}{0.5pt}
       \multirow{2}*{ContiVAE} & \scalebox{0.8}{$\sqrt{\text{MISE}}$}
                           & \textbf{2.5800 $\pm$ 1.48}
                           & \textbf{0.8024 $\pm$ 0.01}
                           & \textbf{0.1005 $\pm$ 0.004}
                           & \textbf{1.5718 $\pm$ 0.73} \\
                           & \scalebox{0.8}{$\sqrt{\text{DPE}}$}
                           & \textbf{4.2986 $\pm$ 3.74}
                           & \textbf{0.1367 $\pm$ 0.02}
                           & \textbf{0.0049 $\pm$ 0.001}
                           & \textbf{0.3196 $\pm$ 0.12} \\
    \specialrule{0.8pt}{0.5pt}{0.5pt}
        \multirow{2}*{MLP} & \scalebox{0.8}{$\sqrt{\text{MISE}}$}
                           & $3.4804 \pm 0.17$
                           & $3.9918 \pm 0.15$
                           & $0.2826 \pm 0.01$
                           & $7.9010\pm 1.04$ \\
                           & \scalebox{0.8}{$\sqrt{\text{DPE}}$}
                           & $7.8602 \pm 0.01$
                           & $10.3197 \pm 0.01$
                           & $0.4665 \pm 0.09$
                           & $8.5948\pm 0.15$ \\
    \specialrule{1.3pt}{0pt}{2pt}
    \end{tabular}
    \label{tab:appendixmlp}
    \caption{Performance of ContiVAE and MLP on the News dataset with reduced training sample size.}
\end{table}

\end{subappendices}

%% file: main.bbl
\begin{thebibliography}{10}
\providecommand{\url}[1]{\texttt{#1}}
\providecommand{\urlprefix}{URL }
\providecommand{\doi}[1]{https://doi.org/#1}

\bibitem{al2015volunteerimpact}
Al~Mutawa, O.: Impact of Volunteer Management Practice on Volunteer Motivation and Satisfaction to Enhance Volunteer Retention. Ph.D. thesis, Brunel University London (2015)

\bibitem{bang2005doubly}
Bang, H., Robins, J.M.: Doubly robust estimation in missing data and causal inference models. Biometrics  \textbf{61}(4),  962--973 (2005)

\bibitem{scigan2020}
Bica, I., Jordon, J., van~der Schaar, M.: Estimating the effects of continuous-valued interventions using generative adversarial networks. In: Proc. of NeurIPS. pp. 16434--16445 (2020)

\bibitem{burt}
Burt, R.S.: Structural holes: The Social Structure of Competition. Harvard University Press, Cambridge, Mass (1992)

\bibitem{cuskelly2006volunteermanagement}
Cuskelly, G., Taylor, T., Hoye, R., Darcy, S.: Volunteer management practices and volunteer retention: A human resource management approach. Sport Management Review  \textbf{9}(2),  141--163 (2006)

\bibitem{floto2022tilted}
Floto, G., Kremer, S., Nica, M.: The tilted variational autoencoder: Improving out-of-distribution detection. In: Proc. of ICLR (2022)

\bibitem{fong2018covariateCBGPS}
Fong, C., Hazlett, C., Imai, K.: Covariate balancing propensity score for a continuous treatment: Application to the efficacy of political advertisements. The Annals of Applied Statistics  \textbf{12}(1),  156--177 (2018)

\bibitem{guo2020survey}
Guo, R., Cheng, L., Li, J., Hahn, P.R., Liu, H.: A survey of learning causality with data: Problems and methods. ACM Computing Surveys (CSUR)  \textbf{53}(4),  1--37 (2020)

\bibitem{imai2014covariateCBPS}
Imai, K., Ratkovic, M.: Covariate balancing propensity score. Journal of the Royal Statistical Society Series B: Statistical Methodology  \textbf{76}(1),  243--263 (2014)

\bibitem{imai2004causalgps}
Imai, K., Van~Dyk, D.A.: Causal inference with general treatment regimes: Generalizing the propensity score. Journal of the American Statistical Association  \textbf{99}(467),  854--866 (2004)

\bibitem{kennedy2017nondoublyrobust}
Kennedy, E.H., Ma, Z., McHugh, M.D., Small, D.S.: Non-parametric methods for doubly robust estimation of continuous treatment effects. Journal of the Royal Statistical Society Series B: Statistical Methodology  \textbf{79}(4),  1229--1245 (2017)

\bibitem{kingma2014adam}
Kingma, D.P., Ba, J.: Adam: A method for stochastic optimization. arXiv preprint arXiv:1412.6980  (2014)

\bibitem{kingma2013autonormalvae}
Kingma, D.P., Welling, M.: Auto-encoding variational bayes. arXiv preprint arXiv:1312.6114  (2013)

\bibitem{volunteer2003}
Locke, M., Ellis, A., Smith, J.D.: Hold on to what you’ve got: the volunteer retention literature. Voluntary Action  \textbf{5}(3),  81--99 (2003)

\bibitem{cevae17}
Louizos, C., Shalit, U., Mooij, J.M., Sontag, D., Zemel, R.S., Welling, M.: Causal effect inference with deep latent-variable models. In: Proc. of NeurIPS (2017)

\bibitem{cider2022}
Ma, J., Dong, Y., Huang, Z., Mietchen, D., Li, J.: Assessing the causal impact of covid-19 related policies on outbreak dynamics: A case study in the us. In: Proc. of ACM Web Conference. pp. 2678--2686 (2022)

\bibitem{montgomery2000measuringproxies}
Montgomery, M.R., Gragnolati, M., Burke, K.A., Paredes, E.: Measuring living standards with proxy variables. Demography  \textbf{37}(2),  155--174 (2000)

\bibitem{pearl2012measurementpearlproxies}
Pearl, J.: On measurement bias in causal inference. arXiv preprint arXiv:1203.3504  (2012)

\bibitem{pearl2016causalprimer}
Pearl, J., Glymour, M., Jewell, N.P.: Causal Inference in Statistics: A Primer (2016)

\bibitem{rezende2014stochasticvae}
Rezende, D.J., Mohamed, S., Wierstra, D.: Stochastic backpropagation and approximate inference in deep generative models. In: Proc. of ICML. pp. 1278--1286 (2014)

\bibitem{rissanen2021criticalcevaefault}
Rissanen, S., Marttinen, P.: A critical look at the consistency of causal estimation with deep latent variable models. Proc. of NeurIPS  \textbf{34},  4207--4217 (2021)

\bibitem{rosenbaum1987ipw}
Rosenbaum, P.R.: Model-based direct adjustment. Journal of the American statistical Association  \textbf{82}(398),  387--394 (1987)

\bibitem{rubin1984bayesianlypotentialoutcome}
Rubin, D.B.: Bayesianly justifiable and relevant frequency calculations for the applied statistician. The Annals of Statistics pp. 1151--1172 (1984)

\bibitem{schwab2020learningdrnet}
Schwab, P., Linhardt, L., Bauer, S., Buhmann, J.M., Karlen, W.: Learning counterfactual representations for estimating individual dose-response curves. In: Proc. of AAAI. vol.~34, pp. 5612--5619 (2020)

\bibitem{tarnet}
Shalit, U., Johansson, F.D., Sontag, D.: Estimating individual treatment tffect: Generalization bounds and algorithms. In: Proc. of ACM ICML. pp. 3076--3085 (2017)

\bibitem{weinstein2013cancertcga}
Weinstein, J.N., Collisson, E.A., Mills, G.B., Shaw, K.R., Ozenberger, B.A., Ellrott, K., Shmulevich, I., Sander, C., Stuart, J.M.: The cancer genome atlas pan-cancer analysis project. Nature Genetics  \textbf{45}(10),  1113--1120 (2013)

\bibitem{yao2021survey}
Yao, L., Chu, Z., Li, S., Li, Y., Gao, J., Zhang, A.: A survey on causal inference. ACM Transactions on Knowledge Discovery from Data (TKDD)  \textbf{15}(5),  1--46 (2021)

\end{thebibliography}
